\documentclass[10pt,twocolumn,letterpaper]{article}

\usepackage[pagenumbers]{wacv} 

\usepackage{algorithm}
\usepackage{algpseudocode}
\usepackage{multirow}
\usepackage{placeins}

\definecolor{wacvblue}{rgb}{0.21,0.49,0.74}
\usepackage[breaklinks,colorlinks,allcolors=wacvblue]{hyperref}

\title{Channel-Wise and Token-Aware Post-Training Quantization for Visual State Space Duality}

\author{
Jonghyeon Lim \qquad Changhoon Yim\\
Intelligent Image Processing Laboratory\\
Department of Computer Science and Engineering, Konkuk University\\
Seoul, Republic of Korea\\
{\tt\small flawhdgus@konkuk.ac.kr, cyim@konkuk.ac.kr}
}

\begin{document}
\maketitle
\begin{abstract}
State space models (SSMs), particularly Mamba, have emerged as efficient alternatives to attention-based architectures and have been extended to vision through ViM, VMamba, and Visual State Space Duality (VSSD). Yet the low-bit post-training quantization (PTQ) behavior of VSSD remains insufficiently understood. A weight--activation split on VSSD-Tiny identifies activation quantization as the dominant low-bit bottleneck, while representative inputs to selected VSSD-backbone linear layers exhibit strong channel-wise magnitude variation and token-localized extremes. We propose the Channel-wise Token-balanced Output-Aware Clipping (CTOAC) method, which learns per-input-channel clipping bounds by minimizing a token-balanced reconstruction loss on the corresponding linear outputs. Only the selected linear layers and their input activations are quantized; other backbone operations retain their original precision. Across VSSD-Tiny, VSSD-Small, and VSSD-Base, the proposed CTOAC method retains ImageNet-1K accuracy and remains substantially more robust than the evaluated baselines at more aggressive precision settings. Applying the same quantization scope to VSSD backbones on COCO and ADE20K preserves strong object detection, instance segmentation, and semantic segmentation performance. An optimized RTX 4090 deployment configuration achieves up to $1.42\times$ end-to-end speedup over FP32.
\end{abstract}
    
\section{Introduction}
\label{sec:introduction}

\begin{figure}[t]
    \centering
    \includegraphics[width=\linewidth]{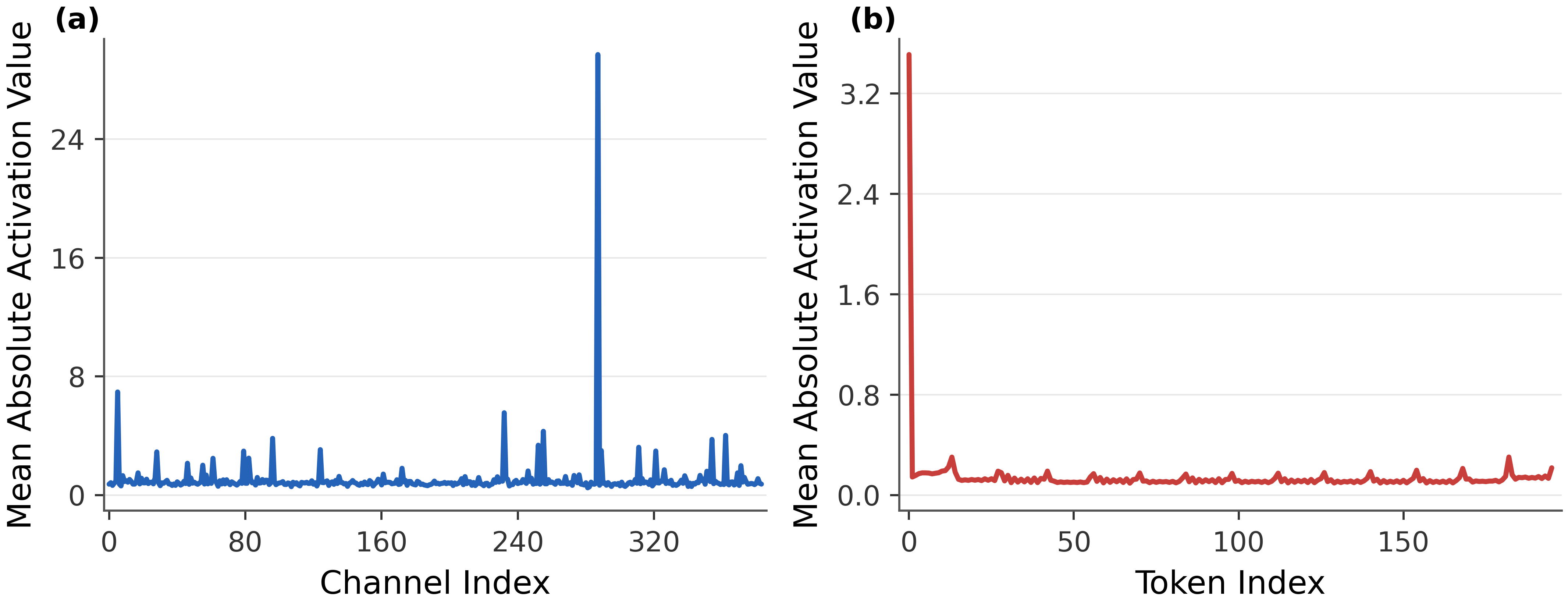}
    \caption{Activation analysis of representative linear inputs in late Stage~3 of VSSD-Tiny. \textbf{(a)} Mean absolute input activation across channels for the input-projection layer in the eighth VSSD block. \textbf{(b)} Mean absolute input activation across tokens for the output-projection layer in the seventh VSSD block.}
    \label{fig:activation_distribution}
\end{figure}

Vision Transformers such as ViT~\cite{dosovitskiy2021vit} and Swin Transformer~\cite{liu2021swin} provide strong visual representations, but global self-attention scales quadratically with token count. State space models offer a more efficient sequence-modeling alternative and have recently been extended to vision through ViM~\cite{zhu2024vim}, VMamba~\cite{liu2024vmamba}, and VSSD~\cite{shi2025vssd}. Among these models, VSSD combines hierarchical visual representation learning with non-causal state space duality and achieves strong performance on classification and dense-prediction tasks, making it an important target for efficient low-bit deployment.

Post-training quantization converts pretrained weights and activations to low-bit representations using a small unlabeled calibration set~\cite{banner2019ptq,li2021brecq,nagel2020adaround}, but its behavior is architecture dependent. On VSSD-Tiny, our weight--activation split retains 83.0\% Top-1 accuracy with W4A32 from an FP32 baseline of 83.7\%, whereas W32A4 falls to 2.4\%. This asymmetry identifies activation quantization, rather than weight quantization, as the primary low-bit bottleneck.

\Cref{fig:activation_distribution} indicates that this bottleneck is associated with heterogeneous channel ranges and token-localized activation extremes. Input magnitudes vary markedly across channels, while a small subset of token positions exhibits localized peaks. The proposed Channel-wise Token-balanced Output-Aware Clipping (CTOAC) method addresses the channel axis by learning independent clipping bounds and the token axis by normalizing linear-output reconstruction errors before aggregation. The optimized clipping bounds and activation quantization parameters are fixed for static inference.

Our contributions are summarized as follows:
\begin{itemize}
    \item We identify activation quantization as the primary low-bit bottleneck in VSSD and characterize channel-wise magnitude variation and token-localized activation tails.
    \item We propose CTOAC, which combines learnable channel-wise clipping with token-balanced linear-output reconstruction while retaining fixed quantization parameters at inference time.
    \item We validate CTOAC across three VSSD backbones, five precision settings, progressive clipping calibration, downstream dense-prediction tasks, and optimized GPU deployment.
\end{itemize}

\FloatBarrier

\section{Related work}
\label{sec:related_work}

\subsection{State space and visual state space models}

Structured state space models represent sequences through recurrent latent-state updates while permitting efficient parallel computation. S4 made this formulation practical for long-sequence modeling through a structured parameterization of the state space operator~\cite{gu2022s4}. Mamba introduced input-dependent selective state transitions, allowing the model to control which information is retained or discarded according to the current input~\cite{gu2024mamba}. State Space Duality subsequently established a connection between structured state space computations and matrix transformations related to attention~\cite{dao2024ssd}.

Several architectures have adapted these ideas to visual recognition. ViM processes image tokens using bidirectional Mamba blocks~\cite{zhu2024vim}. VMamba introduces a hierarchical architecture with multidirectional selective scans~\cite{liu2024vmamba}. LocalMamba studies localized scanning windows for visual state space modeling~\cite{huang2025localmamba}. Visual State Space Duality instead introduces non-causal state space duality and demonstrates competitive performance on image classification, object detection, instance segmentation, and semantic segmentation~\cite{shi2025vssd}. Our work studies the low-bit quantization behavior of VSSD and its linear-input activations.

\subsection{Post-training quantization}

Post-training quantization converts pretrained models to low-bit representations using a small calibration set without end-to-end retraining. Early work established integer-only inference and low-bit CNN quantization~\cite{jacob2018quantization,banner2019ptq}, while AdaRound and BRECQ improved reconstruction by optimizing weight rounding or block outputs~\cite{nagel2020adaround,li2021brecq}. SmoothQuant instead reduces activation quantization difficulty through channel-wise rescaling between activations and weights~\cite{xiao2023smoothquant}.

For Vision Transformers, PTQ methods address attention-specific outputs, non-uniform activations, and inter-channel variation. VT-PTQ, APQ-ViT, and PTQ4ViT introduce attention-aware calibration or scale selection~\cite{liu2021vtptq,ding2022apqvit,yuan2022ptq4vit}, whereas FQ-ViT and AdaLog modify quantization formats for difficult activation distributions~\cite{lin2022fqvit,wu2024adalog}. RepQ-ViT, NoisyQuant, IGQ-ViT, OASQ, ERQ, and DopQ-ViT use reparameterization, grouping, perturbation, or outlier-aware treatment to improve activation quantization~\cite{li2023repqvit,liu2023noisyquant,moon2024igqvit,ma2024oasq,zhong2024erq,yang2024dopqvit}.

Recent reconstruction methods further approximate output sensitivity through Hessian- or Fisher-related objectives, including APHQ-ViT, FIMA-Q, and LS-ViT~\cite{wu2025aphqvit,wu2025fimaq,hwang2026lsvit}. CTOAC differs from these approaches: it does not estimate second-order information or optimize weight rounding, but directly learns channel-wise activation clipping bounds through token-balanced reconstruction of linear outputs.

\subsection{Quantization of Mamba-based models}

PTQ4VM studies post-training quantization for Visual Mamba models, including ViM and VMamba, and introduces Per-Token Static quantization together with Joint Learning of Smoothing Scale and Step Size~\cite{cho2025ptq4vm}. MambaQuant applies variance-aligned rotations to quantize models in the Mamba family~\cite{xu2025mambaquant}. Quamba provides a PTQ recipe for selective state space models~\cite{chiang2025quamba}, whereas SSDi8 focuses on 8-bit quantization for SSD architectures~\cite{kim2026ssdi8}.

Recent work has also explored visual state space quantization from different directions. QMamba targets PTQ for vision state space models~\cite{li2025qmamba}. K-scaled quantization combines scaling and reparameterization for Vision Mamba~\cite{shi2025kscaled}. ViM-VQ applies post-training vector quantization to Visual Mamba~\cite{deng2025vimvq}, while OuroMamba introduces a data-free quantization framework~\cite{ramachandran2025ouromamba}.

Despite this progress, the low-bit activation behavior of VSSD remains insufficiently characterized. These methods primarily focus on smoothing, rotation, quantizer design, or model-specific reconstruction. We instead study the interaction between channel-dependent activation ranges and token-dependent reconstruction weighting in non-causal VSSD linear layers.

\begin{figure*}[t]
    \centering
    \includegraphics[width=\textwidth]{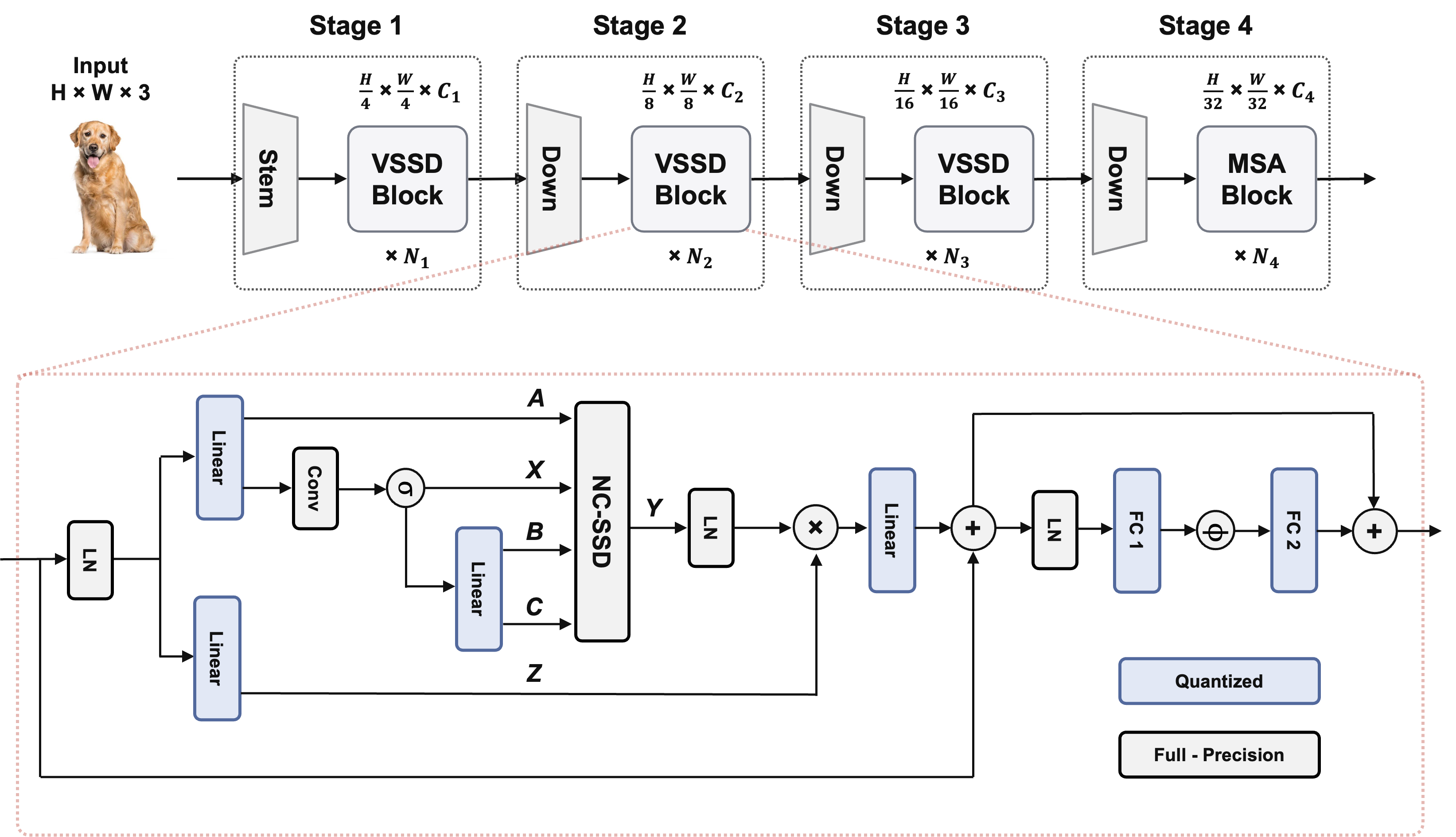}
    \caption{Overview of the hierarchical VSSD architecture and a VSSD block. Highlighted boxes denote the backbone linear layers and their direct input activations quantized by CTOAC; the remaining operations retain their original precision.}
    \label{fig:vssd_architecture}
\end{figure*}

\section{Quantization challenges in VSSD}
\label{sec:quantization_challenges}

VSSD is a hierarchical visual backbone composed of successive spatial stages and non-causal state space duality blocks~\cite{shi2025vssd}. \Cref{fig:vssd_architecture} shows the hierarchical VSSD architecture and the internal structure of a VSSD block. We quantize the weights of the VSSD-backbone linear layers represented by the blue blocks and the activations directly entering them. All remaining backbone operations, including state space computations, convolutions, normalization, nonlinearities, residual paths, and softmax, retain their original full precision.

\subsection{Activation quantization as the primary bottleneck}
\label{sec:activation_bottleneck}

We first isolate the effects of weight and activation quantization using VSSD-Tiny with MinMax calibration. \Cref{tab:weight_activation_split} compares FP32 inference with weight-only W4A32 and activation-only W32A4 under the same linear-layer quantization scope.

\begin{table}[ht]
    \centering
    \small
    \renewcommand{\arraystretch}{1.08}
    \setlength{\tabcolsep}{6pt}
    \begin{tabular}{@{}lrr@{}}
        \toprule
        Bit-width & Top-1 & Top-5 \\
        \midrule
        FP32  & 83.7 & 96.8 \\
        W4A32 & 83.0 & 96.5 \\
        W32A4 &  2.4 &  9.7 \\
        \bottomrule
    \end{tabular}
    \caption{Weight--activation split on VSSD-Tiny using MinMax calibration. Activation quantization is the primary low-bit bottleneck.}
    \label{tab:weight_activation_split}
\end{table}

Reducing only the weights to 4 bits lowers Top-1 accuracy by 0.7 percentage points and Top-5 accuracy by 0.3 percentage points. In contrast, quantizing only the activations to 4 bits reduces Top-1 accuracy by 81.3 percentage points and Top-5 accuracy by 87.1 percentage points. Under the evaluated scope, VSSD weights remain comparatively robust under signed symmetric per-output-channel quantization, whereas 4-bit activations are highly sensitive to range selection.

This asymmetry motivates fixing the quantized weights before calibration and allocating the optimization to activation clipping. The next section examines the activation structure that makes a shared low-bit range ineffective.

\subsection{Channel-wise magnitude variation and token-localized extremes}
\label{sec:activation_variation}

For \cref{fig:activation_distribution}(a), we average absolute full-precision linear inputs over the batch and token dimensions for each input channel. For \cref{fig:activation_distribution}(b), we average over the batch and channel dimensions for each token position. The channel and token views use representative layers to analyze the two axes independently. These statistics are diagnostic only; CTOAC is applied uniformly to every target linear layer and does not depend on selecting these examples during calibration or inference.

\Cref{fig:activation_distribution}(a) shows that a small subset of channels has substantially larger magnitude than the rest, so a shared activation range either wastes resolution on lower-magnitude channels or clips larger channels too aggressively. \Cref{fig:activation_distribution}(b) shows localized token peaks. This does not require a dynamic per-token quantizer at inference; it indicates that an unnormalized reconstruction loss would let high-energy tokens dominate calibration.

\begin{figure}[ht]
    \centering
    \includegraphics[width=\linewidth]{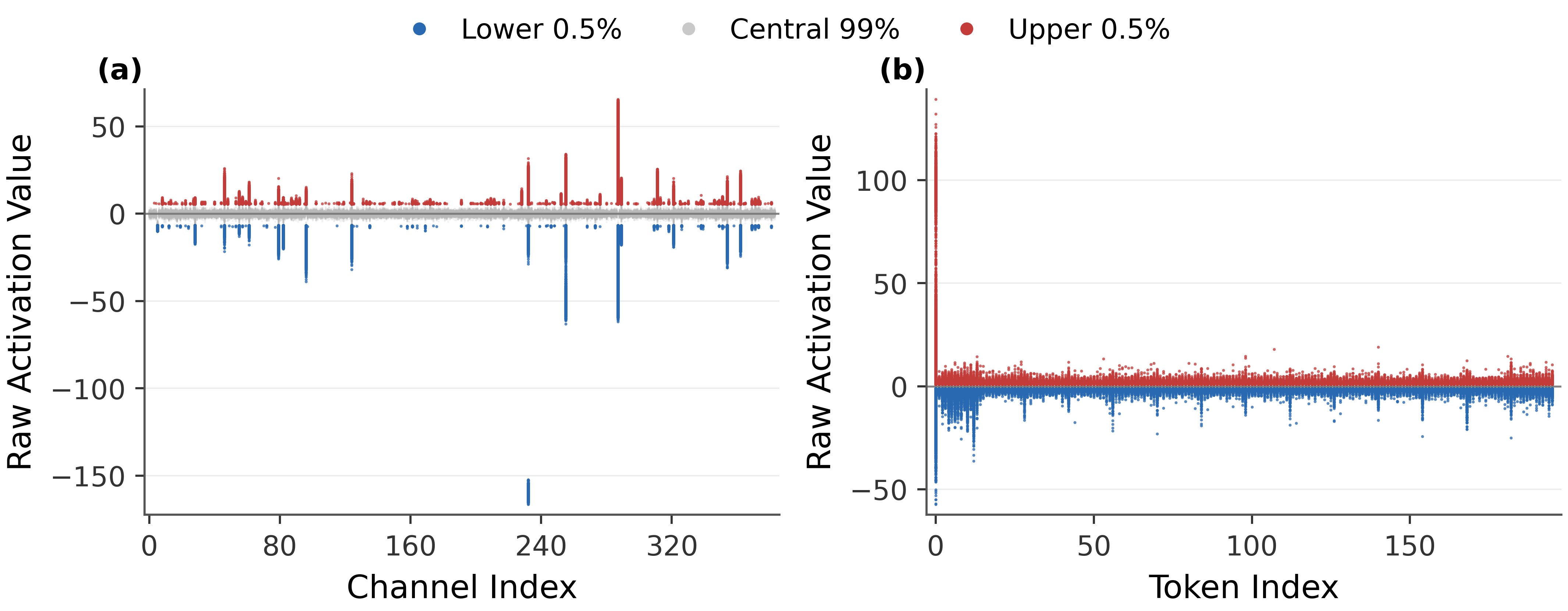}
    \caption{Raw activation tails of representative VSSD-Tiny linear inputs. Blue and red points denote the lower and upper 0.5\% of activation values, while gray points denote the central 99\%. \textbf{(a)} Across input channels, extreme positive and negative values are concentrated in a small subset of channels. \textbf{(b)} Across tokens, extreme responses are concentrated at a small subset of token positions.}
    \label{fig:activation_tails}
\end{figure}

Whereas \cref{fig:activation_distribution} summarizes average magnitude variation, \cref{fig:activation_tails} reveals that the observed heterogeneity is accompanied by localized raw-value tails rather than a uniform shift of the entire distribution. The central 99\% of values remains near zero, while extreme positive and negative responses are concentrated in a small number of channels and token positions. Together, these observations motivate channel-wise clipping bounds and token-normalized output reconstruction.

\FloatBarrier

\section{CTOAC}
\label{sec:method}

\begin{figure*}[t]
    \centering
    \includegraphics[width=\textwidth]{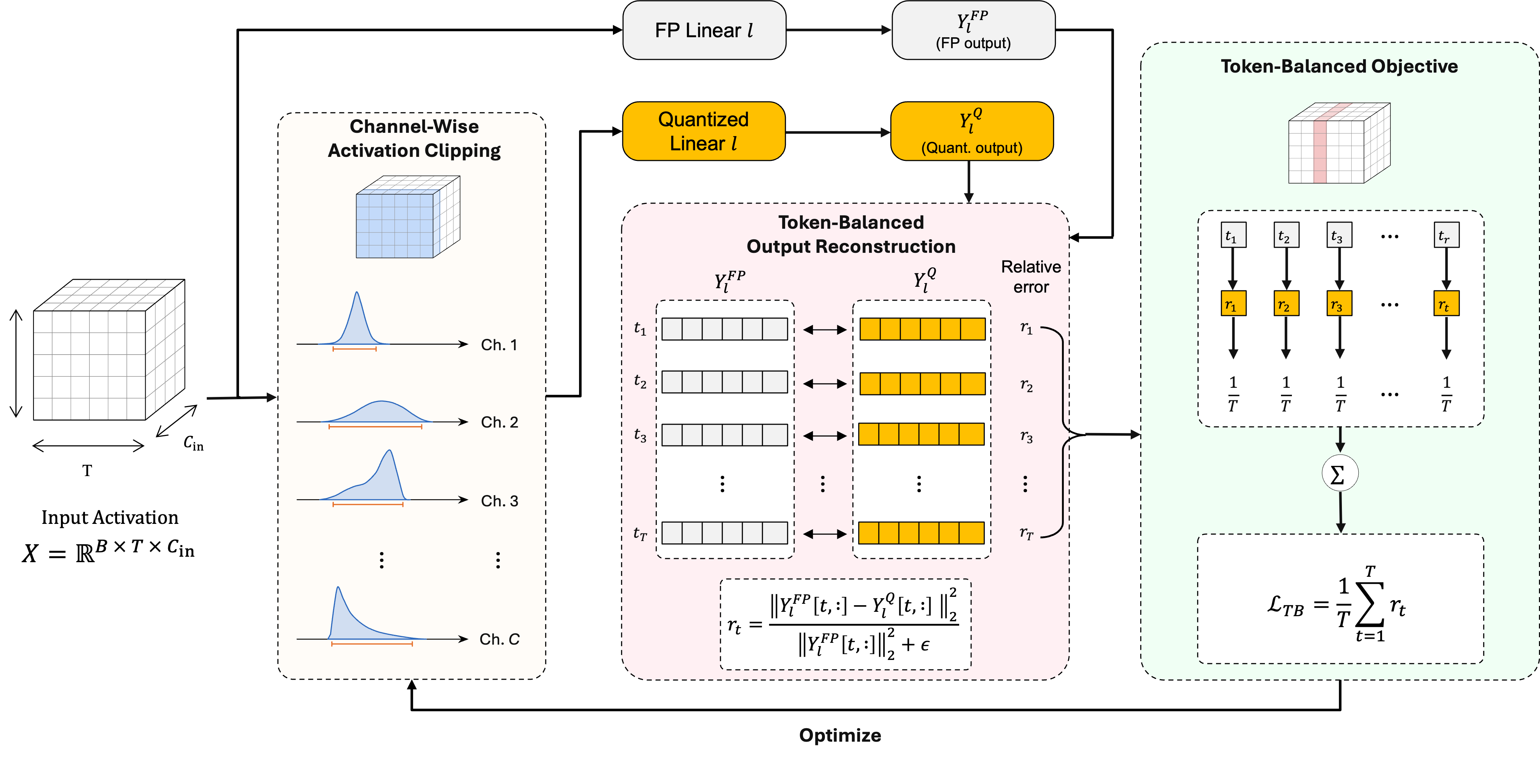}
    \caption{Overview of CTOAC for a target linear layer. Each input channel has independent lower and upper clipping bounds. The full-precision and quantized outputs are compared at every token position, and their relative errors are averaged uniformly to optimize the clipping bounds. The optimized bounds and quantization parameters are fixed after calibration.}
    \label{fig:ctoac_overview}
\end{figure*}

\Cref{fig:ctoac_overview} illustrates the overview of CTOAC method for a target linear layer. The FP branch provides a reference output, while the quantized branch applies channel-wise activation clipping, static activation quantization, and fixed low-bit weights. Their outputs are compared at each token position, and the resulting token-normalized errors are averaged to update only the clipping bounds.

\subsection{Quantization formulation}

Consider a linear layer $l$ with input $X_l$ and weight $W_l$:
\begin{equation}
    X_l \in \mathbb{R}^{B \times T \times C_{\mathrm{in}}},
    \label{eq:input_shape}
\end{equation}
\begin{equation}
    W_l \in \mathbb{R}^{C_{\mathrm{out}} \times C_{\mathrm{in}}}.
    \label{eq:weight_shape}
\end{equation}
The layer also has bias $b_l$. Here, $B$, $T$, $C_{\mathrm{in}}$, and $C_{\mathrm{out}}$ denote the batch size, number of tokens, input channels, and output channels, respectively.

Given scale $s$, zero point $z$, and integer range $[q_{\min}, q_{\max}]$, we define the quantization operator used during calibration as
\begin{equation}
    Q(x; s, z)
    =
    s\left[
        \operatorname{clamp}\!\left(
            \left\lfloor \frac{x}{s} \right\rceil + z,
            q_{\min}, q_{\max}
        \right)-z
    \right].
    \label{eq:affine_quantizer}
\end{equation}
The input is rounded and clamped to the quantization range and then rescaled by $s$ for reconstruction.
During calibration, gradients through the rounding operation are approximated using the straight-through estimator (STE)~\cite{bengio2013estimating}.

Let $Q_W(\cdot)$ denote signed symmetric per-output-channel weight quantization. Activations use static affine asymmetric quantization. Biases and accumulations remain in high precision. The pretrained weight is quantized before activation calibration and then fixed:
\begin{equation}
    W_l^Q = Q_W(W_l).
    \label{eq:weight_quant}
\end{equation}
The quantized weight $W_l^Q$ is fixed throughout calibration. The calibration stage does not optimize weight rounding, pretrained model parameters, or a task-level loss; only the channel-wise activation clipping bounds are updated.

\subsection{Channel-wise activation clipping}

For every input channel $c \in \{1, \ldots, C_{\mathrm{in}}\}$, CTOAC introduces a lower clipping bound $\alpha_{l,c}$ and upper clipping bound $\beta_{l,c}$. For all batch indices $n$ and token positions $t$, the clipped activation is
\begin{equation}
    \overline{X}_{l,n,t,c}
    =
    \operatorname{clip}\!\left(
        X_{l,n,t,c},\,
        \alpha_{l,c},\,
        \beta_{l,c}
    \right).
    \label{eq:channel_clipping}
\end{equation}
The bounds vary across input channels but are shared over all calibration samples and token positions. They are initialized from calibration activations and optimized independently.

Let $\mathcal{C}_A$ denote static affine calibration of the activation quantizer at bit-width $b_a$. The shared scale and zero point for layer $l$ are obtained as
\begin{equation}
    (s_l,z_l)
    =
    \mathcal{C}_A\!\left(\overline{X}_l;b_a\right).
    \label{eq:activation_calibration}
\end{equation}
The clipped activation is then quantized using these parameters:
\begin{equation}
    X_l^Q
    =
    Q\!\left(\overline{X}_l;s_l,z_l\right).
    \label{eq:activation_quant}
\end{equation}
Although the clipping bounds are channel dependent, $s_l$ and $z_l$ are shared across the complete input of layer $l$. CTOAC therefore performs per-channel clipping followed by shared per-layer activation quantization rather than per-channel activation quantization. The optimized bounds and the resulting static activation quantization parameters are fixed after calibration.

\subsection{Token-balanced linear-output reconstruction}

The full-precision linear output is obtained as
\begin{equation}
    Y_l^{\mathrm{FP}}
    =
    X_l \cdot W_l^{\top} + b_l.
    \label{eq:fp_output}
\end{equation}
The corresponding quantized output is obtained as
\begin{equation}
    Y_l^Q
    =
    X_l^Q \cdot (W_l^Q)^{\top} + b_l.
    \label{eq:q_output}
\end{equation}
The dot operator ($\cdot$) denotes matrix multiplication over the input-channel dimension.
CTOAC optimizes the clipping bounds by preserving the output of the linear transformation rather than directly minimizing the difference between full-precision and quantized activation tensors.

For token position $t$, the batch-averaged relative output error is
\begin{equation}
    r_t
    =
    \frac{1}{B}
    \sum_{n=1}^{B}
    \frac{
        \left\|Y_l^{\mathrm{FP}}[n,t,:]-Y_l^Q[n,t,:]\right\|_2^2
    }{
        \left\|Y_l^{\mathrm{FP}}[n,t,:]\right\|_2^2+\epsilon
    }.
    \label{eq:token_error}
\end{equation}
Here, $\epsilon$ is a small constant used for numerical stability.

The token-balanced loss is defined as
\begin{equation}
    \mathcal{L}_{\mathrm{TB}}
    =
    \frac{1}{T}
    \sum_{t=1}^{T} r_t,
    \label{eq:token_balanced_loss_short}
\end{equation}
which is equivalently represented as
\begin{equation}
    \mathcal{L}_{\mathrm{TB}}
    =
    \frac{1}{BT}
    \sum_{n=1}^{B}
    \sum_{t=1}^{T}
    \frac{
        \left\|Y_l^{\mathrm{FP}}[n,t,:]-Y_l^Q[n,t,:]\right\|_2^2
    }{
        \left\|Y_l^{\mathrm{FP}}[n,t,:]\right\|_2^2+\epsilon
    }.
    \label{eq:token_balanced_loss}
\end{equation}

Because normalization precedes token averaging, each token contributes equally in relative-error terms; pooling before normalization would instead overweight high-energy tokens. We refer to $\mathcal{L}_{\mathrm{TB}}$ as token-balanced normalized mean-squared error (TB-NMSE). No separate global reconstruction term is used.

Let
\begin{equation}
    \alpha_l
    =
    [\alpha_{l,1}, \ldots, \alpha_{l,C_{\mathrm{in}}}]
    \label{eq:alpha_vector}
\end{equation}
and
\begin{equation}
    \beta_l
    =
    [\beta_{l,1}, \ldots, \beta_{l,C_{\mathrm{in}}}]
    \label{eq:beta_vector}
\end{equation}
denote the vectors of lower and upper clipping bounds. The calibration problem is
\begin{equation}
    (\alpha_l^{\star}, \beta_l^{\star})
    =
    \operatorname*{arg\,min}_{\alpha_l,\beta_l}
    \;\mathcal{L}_{\mathrm{TB}}(\alpha_l,\beta_l).
    \label{eq:bound_optimization}
\end{equation}
\subsection{Overall calibration procedure}

\Cref{alg:ctoac} summarizes the CTOAC procedure for a target linear layer.

\begin{algorithm}[t]
    \caption{CTOAC procedure for a linear layer.}
    \label{alg:ctoac}
    \begin{algorithmic}[1]
        \Require Full-precision weight $W_l$, bias $b_l$, calibration activation $X_l$, activation bit-width $b_a$, and maximum iteration count $K$
        \Ensure Calibrated $W_l^Q$, $\alpha_l^{\star}$, $\beta_l^{\star}$, $s_l$, and $z_l$
        \State $W_l^Q \leftarrow Q_W(W_l)$
        \State $Y_l^{\mathrm{FP}} \gets X_l \cdot W_l^{\top} + b_l$
        \State Initialize $\alpha_l$ and $\beta_l$ from $X_l$
        \For{$k = 1, \ldots, K$}
            \State $\overline{X}_l \leftarrow \operatorname{clip}(X_l;\alpha_l,\beta_l)$ using \cref{eq:channel_clipping}
            \State $(s_l,z_l) \gets \mathcal{C}_A(\overline{X}_l;b_a)$ using \cref{eq:activation_calibration}
            \State $X_l^Q \gets Q(\overline{X}_l;s_l,z_l)$ using \cref{eq:activation_quant}
            \State $Y_l^Q \gets X_l^Q \cdot (W_l^Q)^{\top} + b_l$ using \cref{eq:q_output}
            \State Compute $\mathcal{L}_{\mathrm{TB}}$ using \cref{eq:token_error,eq:token_balanced_loss}
            \State Update $\alpha_l$ and $\beta_l$
        \EndFor
        \State Recompute $(s_l,z_l)$ using the optimized clipping bounds
        \State \Return $W_l^Q$, $\alpha_l^{\star}$, $\beta_l^{\star}$, $s_l$, and $z_l$
    \end{algorithmic}
\end{algorithm}

The procedure is repeated for every target linear layer using the same calibration set. Once calibration is complete, $W_l^Q$, $\alpha_l^{\star}$, $\beta_l^{\star}$, $s_l$, and $z_l$ are fixed for inference. The full-precision reference branch and reconstruction objective are removed, leaving only fixed channel-wise clipping, shared static activation quantization, and the low-bit linear operation.

\begin{table*}[t]
    \centering
    \small
    \renewcommand{\arraystretch}{1.08}
    \setlength{\tabcolsep}{6pt}
    \begin{tabular}{@{}llccccc@{}}
        \toprule
        Model & Method & W8A8 & W6A6 & W4A4 & W4A3 & W3A3 \\
        \midrule
        \multirow{6}{*}{\shortstack[l]{VSSD-Tiny\\FP32 : 83.7}}
          & MinMax      & 83.4 & 79.1 &  2.0 &  0.3 &  0.3 \\
          & Truncation  & 79.5 & 72.2 & 10.2 &  1.1 &  1.6 \\
          & SmoothQuant~\cite{xiao2023smoothquant} & 83.6 & 79.6 &  1.3 &  0.8 &  0.6 \\
          & BRECQ~\cite{li2021brecq}       & 83.6 & 77.9 & 19.1 &  0.1 &  0.1 \\
          & PTQ4VM~\cite{cho2025ptq4vm}      & 83.4 & 83.0 & 81.0 &  6.8 &  3.9 \\
          & \textbf{CTOAC} & \textbf{83.6} & \textbf{83.3} & \textbf{81.1} & \textbf{47.0} & \textbf{26.3} \\
        \midrule
        \multirow{6}{*}{\shortstack[l]{VSSD-Small\\FP32 : 84.6}}
          & MinMax      & 84.5 & 83.1 &  0.6 &  0.3 &  0.3 \\
          & Truncation  & 80.2 & 79.1 &  3.7 &  0.4 &  0.1 \\
          & SmoothQuant~\cite{xiao2023smoothquant} & 84.5 & 83.5 &  1.3 &  0.2 &  0.2 \\
          & BRECQ~\cite{li2021brecq}       & 84.5 & 83.8 &  2.0 &  0.1 &  0.1 \\
          & PTQ4VM~\cite{cho2025ptq4vm}      & 84.2 & 84.0 & 82.3 &  4.7 &  2.8 \\
          & \textbf{CTOAC} & \textbf{84.6} & \textbf{84.5} & \textbf{83.2} & \textbf{41.0} & \textbf{42.2} \\
        \midrule
        \multirow{6}{*}{\shortstack[l]{VSSD-Base\\FP32 : 85.4}}
          & MinMax      & 85.4 & 84.7 &  0.1 &  0.1 &  0.1 \\
          & Truncation  & 83.4 & 81.3 &  1.0 &  0.1 &  0.2 \\
          & SmoothQuant~\cite{xiao2023smoothquant} & 85.3 & 84.8 &  0.6 &  0.1 &  0.1 \\
          & BRECQ~\cite{li2021brecq}       & 85.3 & 84.9 &  0.1 &  0.1 &  0.1 \\
          & PTQ4VM~\cite{cho2025ptq4vm}      & 85.3 & 85.2 & 84.2 &  5.5 &  3.1 \\
          & \textbf{CTOAC} & \textbf{85.4} & \textbf{85.3} & \textbf{84.8} & \textbf{76.9} & \textbf{74.2} \\
        \bottomrule
    \end{tabular}
    \caption{ImageNet-1K Top-1 accuracy (\%) under low-bit quantization. The result of FP32 baseline is shown with each model. The results of the proposed CTOAC method are shown in bold.}
    \label{tab:imagenet_results}
\end{table*}

\section{Experiments}
\label{sec:experiments}

We evaluate VSSD-Tiny, VSSD-Small, and VSSD-Base on ImageNet-1K~\cite{russakovsky2015imagenet}. Classification calibration uses 256 training images, and final accuracy is measured on all 50,000 validation images. We use $WiAj$ to denote $i$-bit weight and $j$-bit activation quantization, and evaluate W8A8, W6A6, W4A4, W4A3, and W3A3 settings.

For ImageNet classification, we compare CTOAC with MinMax, percentile-based Truncation, SmoothQuant~\cite{xiao2023smoothquant}, BRECQ~\cite{li2021brecq}, and PTQ4VM~\cite{cho2025ptq4vm} under the same calibration and evaluation protocol. For downstream tasks, we compare CTOAC with MinMax, Truncation, and SmoothQuant, which are applied consistently within the corresponding detection and segmentation pipelines. For every method, only the selected VSSD-backbone linear layers and their direct input activations are quantized; all other backbone operations retain their original full precision.

For each downstream task, the selected backbone linear layers and their direct input activations are recalibrated using unlabeled images from the corresponding training set, while the task-specific head remains in full precision. For COCO~\cite{lin2014coco}, we evaluate object detection and instance segmentation using box AP $\mathrm{AP}^{b}$ and mask AP $\mathrm{AP}^{m}$, respectively. For ADE20K~\cite{zhou2017ade20k}, we evaluate semantic segmentation using single-scale and multi-scale mIoU.

\subsection{ImageNet classification}

\Cref{tab:imagenet_results} shows that the main difficulty emerges when activation precision enters the 4-bit regime. CTOAC remains within 0.4 percentage points of FP32 at W8A8 and W6A6 across all three backbones, indicating that its calibration does not sacrifice accuracy at moderate precision. At W4A4, CTOAC improves over PTQ4VM by 0.1, 0.9, and 0.6 percentage points on VSSD-Tiny, VSSD-Small, and VSSD-Base, respectively, while the remaining baselines degrade severely.

The separation becomes much larger in the A3 settings. All evaluated baselines fall to single-digit accuracy, whereas CTOAC retains meaningful accuracy across all three models. VSSD-Base is particularly tolerant to aggressive quantization, retaining 76.9\% at W4A3 and 74.2\% at W3A3. These results indicate that the benefit of channel-aware clipping and token-balanced reconstruction becomes more pronounced as the precision setting becomes more aggressive. Each precision setting is calibrated independently, so the resulting accuracies are not constrained to vary monotonically with bit width.

\subsection{Progressive channel-wise clipping calibration}

\Cref{tab:progressive_calibration} progressively introduces the two design axes of CTOAC. The MinMax baseline uses an untrimmed shared activation range. Replacing it with fixed channel-wise clipping substantially improves all three models, confirming that a single range is poorly matched to heterogeneous channels. However, the recovered accuracy remains highly model dependent and far below the final result.

\begin{table}[H]
    \centering
    \small
    \renewcommand{\arraystretch}{1.10}
    \setlength{\tabcolsep}{5pt}
    \begin{tabular}{@{}lccc@{}}
        \toprule
        Model & \shortstack{MinMax\\(No Clipping)} & \shortstack{Fixed Channel-wise\\Clipping} & CTOAC \\
        \midrule
        VSSD-Tiny  & 2.0 & 32.2 & \textbf{81.1} \\
        VSSD-Small & 0.6 & 17.9 & \textbf{83.2} \\
        VSSD-Base  & 0.1 & 60.7 & \textbf{84.8} \\
        \bottomrule
    \end{tabular}
    \caption{Progressive W4A4 calibration analysis on ImageNet-1K. We begin with a shared MinMax activation range, introduce fixed channel-wise clipping, and finally optimize the channel-wise bounds using token-balanced linear-output reconstruction. Each entry reports Top-1 accuracy (\%).}
    \label{tab:progressive_calibration}
\end{table}

The complete CTOAC method further optimizes the channel-wise bounds using TB-NMSE on the corresponding linear outputs, restoring 81.1\%, 83.2\%, and 84.8\% Top-1 accuracy. The progressive calibration analysis shows that static channel separation alone cannot resolve the W4A4 failure; the bounds must be calibrated with respect to their output distortion. This analysis shows that fixed channel-wise clipping alone is insufficient, whereas the complete CTOAC calibration substantially recovers W4A4 accuracy.

\FloatBarrier

\begin{table}[t]
    \centering
    \footnotesize
    \renewcommand{\arraystretch}{1.10}
    \setlength{\tabcolsep}{2pt}
    \begin{tabular}{@{}llccc@{}}
        \toprule
        Model & Method & W8A8 & W6A6 & W4A4 \\
        \midrule
        \multirow{4}{*}{\shortstack[l]{VSSD-Tiny\\FP32 : 47.0 / 42.6}}
          & MinMax      & 46.9 / 42.6 & 46.3 / 42.0 & 0.3 / 0.3 \\
          & Truncation  & 44.5 / 40.5 & 44.1 / 40.0 & 1.6 / 1.6 \\
          & SmoothQuant & 46.9 / 42.6 & 46.5 / 42.1 & 0.6 / 0.6 \\
          & \textbf{CTOAC} & \textbf{47.0 / 42.6} & \textbf{46.8 / 42.6} & \textbf{45.3 / 41.4} \\
        \midrule
        \multirow{4}{*}{\shortstack[l]{VSSD-Small\\FP32 : 48.3 / 43.5}}
          & MinMax      & 48.3 / 43.4 & 47.9 / 43.0 & 0.0 / 0.0 \\
          & Truncation  & 45.6 / 41.0 & 44.9 / 40.4 & 0.3 / 0.3 \\
          & SmoothQuant & 48.2 / 43.4 & 47.8 / 43.0 & 0.1 / 0.1 \\
          & \textbf{CTOAC} & \textbf{48.2 / 43.5} & \textbf{48.1 / 43.4} & \textbf{47.1 / 42.6} \\
        \bottomrule
    \end{tabular}
    \caption{Object detection and instance segmentation on COCO. Each entry reports $\mathrm{AP}^{b}/\mathrm{AP}^{m}$. CTOAC results are highlighted in bold.}
    \label{tab:coco_results}
\end{table}

\subsection{Object detection and instance segmentation}

\Cref{tab:coco_results} presents the object detection and instance segmentation results on COCO.

At W8A8 and W6A6, CTOAC remains within 0.2 AP of the corresponding FP32 models. At W4A4, it limits the degradation to 1.7 / 1.2 box and mask AP on VSSD-Tiny and 1.2 / 0.9 AP on VSSD-Small, whereas the calibration baselines collapse. The similar retention of box and mask AP indicates that the quantized backbone preserves both object-level localization and the spatial detail required for instance-mask prediction. This provides stronger evidence than classification alone because the full-precision downstream heads operate on features produced by the quantized backbone linear layers.

\begin{table}[t]
    \centering
    \footnotesize
    \renewcommand{\arraystretch}{1.10}
    \setlength{\tabcolsep}{2pt}
    \begin{tabular}{@{}llccc@{}}
        \toprule
        Model & Method & W8A8 & W6A6 & W4A4 \\
        \midrule
        \multirow{4}{*}{\shortstack[l]{VSSD-Tiny\\FP32 : 47.8 / 48.7}}
          & MinMax      & 47.8 / 48.6 & 47.3 / 48.1 & 0.9 / 1.0 \\
          & Truncation  & 43.8 / 44.2 & 40.5 / 40.5 & 1.5 / 1.9 \\
          & SmoothQuant & 47.8 / 48.6 & 47.4 / 48.3 & 1.2 / 1.3 \\
          & \textbf{CTOAC} & \textbf{47.8 / 48.6} & \textbf{47.5 / 48.2} & \textbf{45.1 / 46.3} \\
        \bottomrule
    \end{tabular}
    \caption{Semantic segmentation on ADE20K. Each entry reports single-scale / multi-scale mIoU. CTOAC results are highlighted in bold.}
    \label{tab:ade20k_results}
\end{table}

\subsection{Semantic segmentation}

\Cref{tab:ade20k_results} presents the semantic segmentation results on ADE20K.

On ADE20K, CTOAC reduces W4A4 single-scale and multi-scale mIoU by 2.7 and 2.4 points from FP32, respectively, while the other calibration baselines fall near zero. The comparable degradation under single- and multi-scale evaluation indicates that the quantized backbone retains spatially coherent features across evaluation scales.

\begin{figure}[t]
    \centering
    \includegraphics[width=0.90\linewidth]{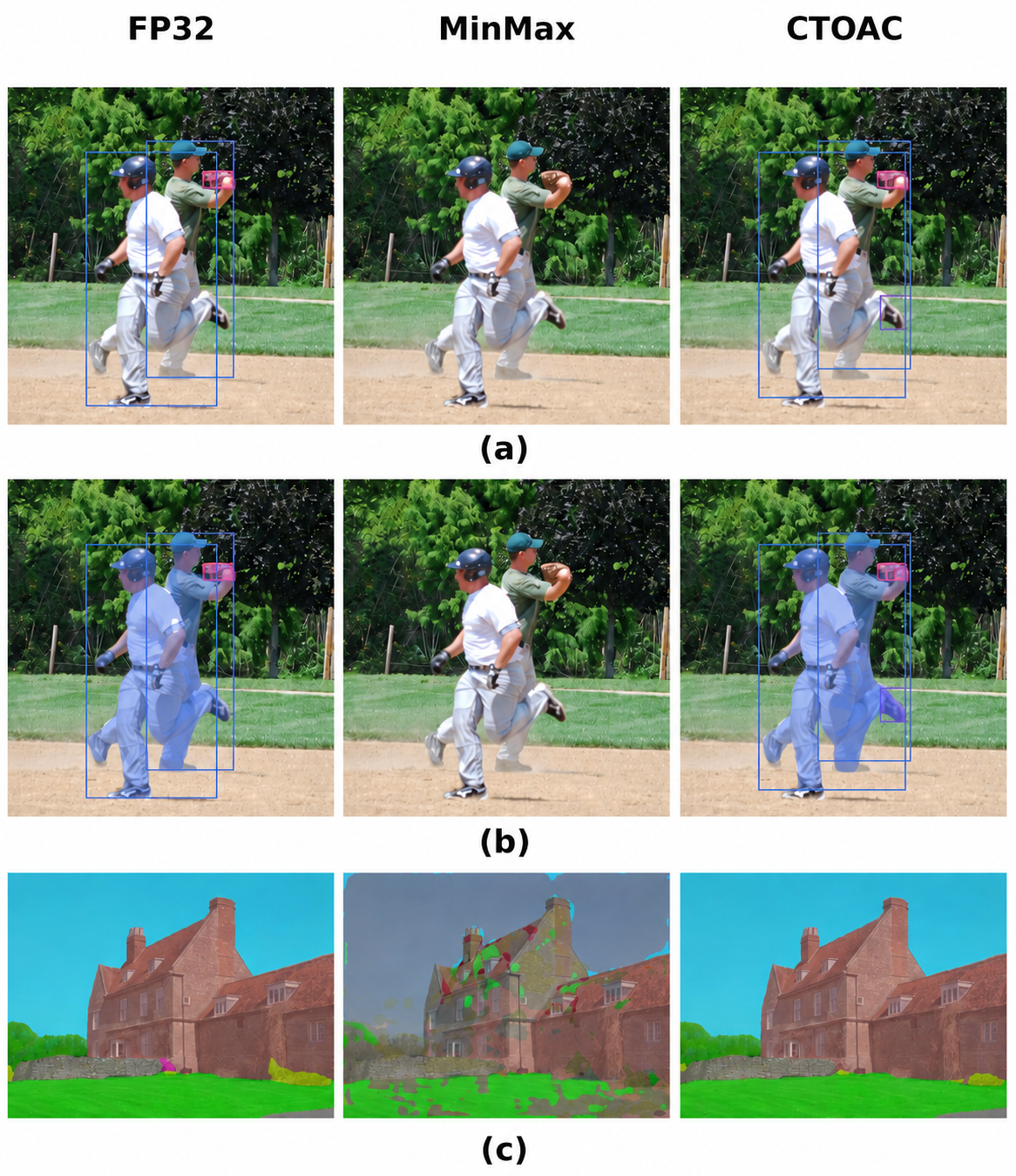}
    \caption{Qualitative comparison for VSSD-Tiny under W4A4 quantization. Columns show FP32, MinMax, and CTOAC. \textbf{(a)} Object detection, \textbf{(b)} Instance segmentation, and \textbf{(c)} Semantic segmentation.}
    \label{fig:downstream_qualitative}
\end{figure}

\Cref{fig:downstream_qualitative} complements \Cref{tab:coco_results,tab:ade20k_results} with qualitative comparisons under identical post-processing and visualization settings. Under W4A4, MinMax loses detections and produces fragmented masks, whereas CTOAC retains object instances, mask boundaries, and semantic regions that remain visually closer to FP32 across all three downstream tasks.

\subsection{Deployment efficiency}

\begin{figure}[t]
    \centering
    \includegraphics[width=0.90\linewidth]{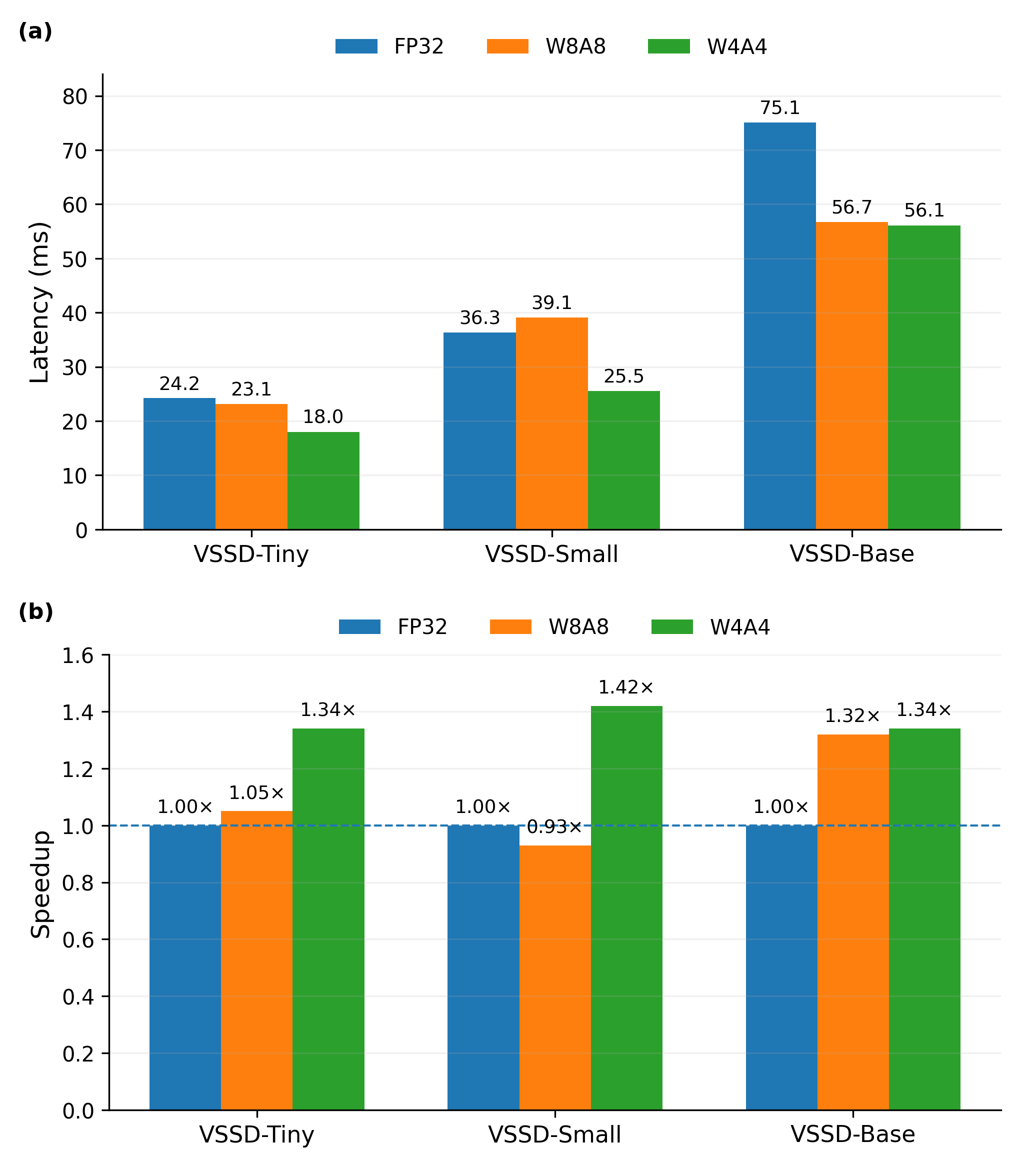}
    \caption{CUTLASS-based deployment efficiency of CTOAC on an RTX 4090 with batch size 32: \textbf{(a)} End-to-end inference latency and \textbf{(b)} Speedup over FP32.}
    \label{fig:deployment_speedup}
\end{figure}

\Cref{fig:deployment_speedup} evaluates CUTLASS-based low-bit deployment on an NVIDIA RTX 4090 at batch size 32. W4A4 provides consistent $1.34\times \sim 1.42\times$ speedup across all three backbones, whereas W8A8 ranges from a slight slowdown to a $1.32\times$ speedup. In particular, the similar W8A8 and W4A4 latency on VSSD-Base shows that end-to-end latency is not determined by arithmetic precision alone. Kernel efficiency, data-layout conversion, launch overhead, and the remaining original-precision operations can dominate the realized latency. Therefore the measurements demonstrate practical gains for the evaluated deployment configurations rather than a universal arithmetic-only 4-bit speedup.

\section{Conclusion}
\label{sec:conclusion}

We proposed the CTOAC method, which is an efficient post-training quantization method for selected VSSD-backbone linear layers and their direct input activations. Motivated by channel-wise magnitude variation and token-localized activation tails, CTOAC learns channel-specific clipping bounds through token-balanced reconstruction of linear outputs. The learned bounds and shared per-layer activation quantizers are fixed after calibration, while all non-target backbone operations retain their original full-precision.

Across VSSD-Tiny, VSSD-Small, and VSSD-Base, CTOAC preserves near-FP32 accuracy at W8A8 and W6A6, strong accuracy at W4A4, and substantially greater robustness than the evaluated baselines at A3 precision. The progressive analysis shows that fixed channel-wise clipping alone is insufficient and that the complete output-aware calibration is required to recover low-bit accuracy. After task-specific recalibration, the same quantization scope preserves features required for object detection and instance segmentation on COCO and semantic segmentation on ADE20K. With CUTLASS-based low-bit kernels, the evaluated W4A4 deployment configurations achieve up to $1.42\times$ end-to-end speedup on an RTX 4090.

These results establish channel-wise range adaptation and token-balanced output reconstruction as an effective calibration strategy for accurate and deployable low-bit VSSD models.

\section*{Acknowledgments}

This work was supported by the National Research Foundation of Korea
(NRF) grant funded by the Korean government (MSIT) (RS-2026-25476371).

\clearpage
{
    \small
    \bibliographystyle{ieeenat_fullname}
    \bibliography{main}
}

\end{document}